\documentclass[letterpaper]{article} % DO NOT CHANGE THIS
\usepackage{aaai2026}    % DO NOT CHANGE THIS
\usepackage{times}                   % DO NOT CHANGE THIS
\usepackage{helvet}                  % DO NOT CHANGE THIS
\usepackage{courier}                 % DO NOT CHANGE THIS
\usepackage[hyphens]{url}            % DO NOT CHANGE THIS
\usepackage{graphicx}                % DO NOT CHANGE THIS
\usepackage{natbib}                  % DO NOT CHANGE THIS AND DO NOT ADD ANY OPTIONS TO IT
\usepackage{caption}                 % DO NOT CHANGE THIS AND DO NOT ADD ANY OPTIONS TO IT
\usepackage{algorithm}
\usepackage{algorithmic}

\usepackage[most]{tcolorbox}
\tcbuselibrary{listingsutf8}
\usepackage{newfloat}
\usepackage{listings}
\lstdefinelanguage{json}{
    basicstyle=\ttfamily\footnotesize,
    stepnumber=1,
    numbersep=5pt,
    showstringspaces=false,
    breaklines=true,
    backgroundcolor=\color{gray!10},
    literate=
     *{0}{{0}}{1}
      {1}{{1}}{1}
      {2}{{2}}{1}
      {3}{{3}}{1}
      {4}{{4}}{1}
      {5}{{5}}{1}
      {6}{{6}}{1}
      {7}{{7}}{1}
      {8}{{8}}{1}
      {9}{{9}}{1}
}
\DeclareCaptionStyle{ruled}{labelfont=normalfont,labelsep=colon,strut=off} % DO NOT CHANGE THIS
\floatstyle{ruled}
\newfloat{listing}{tb}{lst}{}
\floatname{listing}{Listing}

\usepackage{booktabs}       % professional-quality tables
\usepackage{amsfonts}       % blackboard math symbols
\usepackage{nicefrac}       % compact symbols for 1/2, etc.
\usepackage{microtype}      % microtypography
\usepackage{pifont}
\usepackage{amssymb}
\usepackage{makecell}
\usepackage{tabularx}
\usepackage[dvipsnames]{xcolor}
\usepackage{multirow}
\usepackage{bm}
\usepackage{amsmath}
\usepackage[most]{tcolorbox}
\usepackage{array}          % 提供更灵活的表格列格式
\usepackage[table]{xcolor}
\usepackage{tikz}
\usepackage{xcolor}
\usepackage{makecell}

\newtcolorbox{promptbox}[2][]{%
  colback=#2!5!white,
  colframe=#2!75!black,
  width=\textwidth,
  boxrule=0.5mm,
  arc=1mm,
  title=\textbf{#1},
  #1
}
\definecolor{ogreen}{RGB}{34, 139, 34}

% ...

\title{SPAN: Benchmarking and Improving \\Cross-Calendar Temporal Reasoning of Large Language Models}
\author{Zhongjian Miao\textsuperscript{\rm 1}, Hao Fu\textsuperscript{\rm 1}\thanks{~Corresponding Author.}\thanks{~Work done when Hao Fu was at Li Auto.}, Chen Wei\textsuperscript{\rm 1}}
\affiliations{\textsuperscript{\rm 1}Li Auto Inc.\\ \{miaozhongjian, chenwei10\}@lixiang.com}

\usepackage{bibentry}
\begin{document}
\maketitle
\begin{abstract}
Temporal reasoning is a fundamental capability for large language models (LLMs) to understand real-world dynamics. Existing research on temporal reasoning has predominantly focused on the Gregorian calendar. However, as many countries and regions concurrently adopt multiple calendar systems, temporal reasoning across calendars becomes crucial for LLMs in global and multicultural contexts. Unfortunately, cross-calendar temporal reasoning remains underexplored, with no dedicated benchmark available to evaluate this capability.
To bridge this gap, we introduce \textbf{SPAN}, a cro\textbf{S}s-calendar tem\textbf{P}oral re\textbf{A}soning be\textbf{N}chmark, which requires LLMs to perform intra-calendar temporal reasoning and inter-calendar temporal conversion. SPAN features ten cross-calendar temporal reasoning directions, two reasoning types, and two question formats across six calendars. To enable time-variant and contamination-free evaluation, we propose a template-driven protocol for dynamic instance generation that enables assessment on a user-specified Gregorian date. We conduct extensive experiments on both open- and closed-source state-of-the-art~(SOTA) LLMs over a range of dates spanning $100$ years from $1960$ to $2060$. Our evaluations show that these LLMs achieve an average accuracy of only $34.5$\%, with none exceeding $80$\%, indicating that this task remains challenging. Through in-depth analysis of reasoning types, question formats, and temporal reasoning directions, we identify two key obstacles for LLMs: \textit{Future-Date Degradation} and \textit{Calendar Asymmetry Bias}. 
To strengthen LLMs' cross-calendar temporal reasoning capability, we further develop an LLM-powered \texttt{Time}~\texttt{Agent} that leverages tool-augmented code generation. Empirical results show that \texttt{Time}~\texttt{Agent} achieves an average accuracy of $95.31$\%, outperforming several competitive baselines, highlighting the potential of tool-augmented code generation to advance cross-calendar temporal reasoning. We hope this work will inspire further efforts toward more temporally and culturally adaptive LLMs. Our source code and datasets are available at \url{https://github.com/miaozhongjian/span.git}.
\end{abstract}

\section{Introduction}\label{section:instroduction}
Recent advances in large language models (LLMs) have led to substantial progress across a range of reasoning tasks~\cite{qiao-etal-2023-reasoning,xu2024faithful,li2024understanding,fan-etal-2025-slam}. 
Among these, temporal reasoning, as a fundamental capability for LLMs to understand real-world dynamics, has garnered increasing attention, leading to the development of multiple evaluation benchmarks~\cite{chen2021datasetansweringtimesensitivequestions,liska2022streamingqa, kasai2023realtime,tan2023benchmarkingimprovingtemporalreasoning,wei-etal-2023-menatqa,tan-etal-2023-towards,fatemi2024test,wang2024trambenchmarkingtemporalreasoning,chu-etal-2024-timebench, wang-zhao-2024-tram, ge2025tremu, wei2025time, saxena2025lost}. Nevertheless, most existing efforts focus on temporal reasoning within the Gregorian calendar, neglecting the exploration of LLMs' capability to reason across calendar systems. 
In practice, numerous countries and regions employ an alternative calendar system alongside the widely-used Gregorian calendar system~\cite{taqizadeh1939various,graumann2015problem}. For instance, both Saudi Arabia and China use the Gregorian calendar for official purposes, but Saudi Arabia follows the Islamic calendar for religious observances while China relies on the Chinese lunar calendar for traditional cultural activities. These real-world practices underscore the necessity for reasoning and converting dates across calendar systems, \textit{i.e.,} cross-calendar temporal reasoning, which is crucial for global and multicultural applications of LLMs.

Despite its practical significance, the cross-calendar temporal reasoning capability of LLMs remains unexplored. 
Critically, there is currently no benchmark for assessing this capability, which hinders further research progress. Moreover, existing temporal reasoning benchmarks are inherently time-invariant, limiting evaluation to a single, fixed reference date when the benchmark was created. 
This leads to two drawbacks: 
(1) \textbf{\textit{Temporal Scope Limitation.}} Given the dynamic nature of time, robust temporal reasoning evaluation should ensure a broad temporal scope by enabling assessments at arbitrary reference dates, which is not yet supported by existing benchmarks. 
(2) \textbf{\textit{Data Contamination.}} The ground truth in certain temporal reasoning benchmarks is typically deterministic and publicly available through platforms such as Hugging Face and GitHub. Such accessibility may lead to evaluation data being absorbed into training corpora, potentially biasing evaluations~\cite{sainz-etal-2023-nlp,deng-etal-2024-investigating,balloccu2024leak}. 

In response to these challenges, we introduce \textbf{SPAN}, a cro\textbf{S}s-calendar tem\textbf{P}oral re\textbf{A}soning be\textbf{N}chmark. Unlike prior Gregorian-centric temporal reasoning benchmarks, SPAN requires LLMs to perform temporal reasoning within one calendar based on a reference date, and then convert the result to its equivalent in another calendar. To mitigate data contamination and time-invariant problems, we propose a novel evaluation protocol that dynamically generates instances by instantiating a set of question–code template pairs. By utilizing these carefully-designed template pairs, SPAN generates diverse evaluation instances covering three key dimensions: (1)~$\textbf{\textit{Temporal Reasoning Directions}}$. SPAN supports ten cross-calendar reasoning directions between the Gregorian and five other calendars: Chinese Lunar, Shaka, Hebrew, Islamic, and Persian. (2)~$\textbf{\textit{Reasoning Types}}$. SPAN encompasses date- and festival-based reasoning tasks that perform cross-calendar conversions for general dates and festival dates, respectively. (3)~$\textbf{\textit{Question Formats}}$. SPAN comprises both polar and content questions, where the answers to polar questions are binary judgments (\textit{i.e.,} ``Yes'' or ``No''), whereas content questions require calendar-specific dates as answers. During evaluation, these question-code template pairs are instantiated to generate questions and corresponding code snippets, which are then executed to automatically derive the answers. Overall, SPAN offers a dynamic mechanism to generate varied evaluation instances, enabling robust assessment of LLMs' cross-calendar temporal reasoning.

We conduct comprehensive experiments on both open- and closed-source state-of-the-art LLMs over a range of dates spanning a $100$-year period from $1960$ to $2060$. We demonstrate that these LLMs achieve an average accuracy of only $34.5$\% across the evaluation dates, with none exceeding $80$\% accuracy. We also investigate the impact of reasoning types, question formats, and temporal reasoning directions, identifying two key obstacles for LLMs: \textit{Future-Date Degradation}, where LLMs struggle with cross-calendar temporal reasoning under future reference dates, and \textit{Calendar Asymmetry Bias}, an asymmetry where reasoning from the Gregorian calendar is more accurate than the reverse. 
To advance LLMs' cross-calendar temporal reasoning capabilities, we develop an LLM-powered \texttt{Time}~\texttt{Agent} with tool-augmented code generation.
Empirical results demonstrate that our method achieves an average accuracy of $95.31$\%, surpassing competitive baselines. Our contributions are summarized as follows:
\begin{itemize}
    \item We introduce \textbf{SPAN}, a benchmark assessing LLMs' cross-calendar temporal reasoning across ten directions, two reasoning types, and two question formats.
    \item We propose a template-driven protocol for dynamic instance generation, enabling broad temporal evaluation while mitigating data contamination.
    \item We conduct comprehensive experiments and in-depth analyses, revealing that LLMs struggle to reason across calendars and identifying the obstacles of \textit{Future-Date Degradation} and \textit{Calendar Asymmetry Bias}.
    \item We develop an LLM-powered \textit{Time} \textit{Agent} using tool-augmented code generation, which achieves an average accuracy of $95.31$\%, surpassing multiple strong baselines and demonstrating the potential of tool-augmented code generation for the cross-calendar temporal reasoning task.
\end{itemize}
\section{Related Work}
\paragraph{Temporal Reasoning Benchmarks for LLMs.} Temporal reasoning, the capability to perceive and understand the world’s dynamics, is essential for advancing LLMs toward artificial general intelligence (AGI). Early benchmarks such as \textit{TimeQA}~\cite{chen2021datasetansweringtimesensitivequestions} and \textit{TimeDial}~\cite{qin-etal-2021-timedial} concentrate on time-sensitive question answering and temporal commonsense in dialogue contexts. Recently, a new wave of benchmarks has emerged, targeting more fine-grained, complex temporal reasoning scenarios~\cite{chen2021datasetansweringtimesensitivequestions,liska2022streamingqa, kasai2023realtime,tan2023benchmarkingimprovingtemporalreasoning,wei-etal-2023-menatqa,tan-etal-2023-towards,wang2024trambenchmarkingtemporalreasoning,chu-etal-2024-timebench, wang-zhao-2024-tram, fatemi2024test, ge2025tremu, wei2025time, saxena2025lost}. For example, \textit{TimeBench}~\cite{chu-etal-2024-timebench} introduces a hierarchical benchmark for evaluating a wide spectrum of temporal reasoning capabilities, ranging from temporal relation classification to duration estimation and multi-hop time-aware question answering, spanning symbolic, commonsense, and event-centric tasks. \textit{StreamingQA}~\cite{liska2022streamingqa} and \textit{RealtimeQA}~\cite{kasai2023realtime} extend the focus to dynamic settings where new information arrives continuously, and LLMs are expected to revise their responses over time.
Additionally, \textit{Lost in Time}~\cite{saxena2025losttimeclockcalendar} explores how multimodal LLMs process time-related visual inputs, including analogue clocks and yearly calendars. These benchmarks collectively enable systematic evaluation of LLMs' temporal reasoning across diverse tasks, complexities, and modalities.
\paragraph{Improving Temporal Reasoning Capabilities of LLMs.} In pursuit of human-level temporal reasoning, recent research has proposed various strategies to enhance LLMs~\cite{wei-etal-2023-menatqa,tan-etal-2023-towards,yang-etal-2023-upon,jain2023language,sutimo,xiong2024large,yang-etal-2024-enhancing-temporal}. For instance, \textit{RemeMo}~\cite{yang-etal-2023-upon} improves LLMs' temporal understanding during pre-training by organizing events or sentences according to their temporal order, facilitating the model's capture of complex temporal relationships. \textit{TG-LLM}~\cite{xiong2024large} applies supervised fine-tuning on a synthetic temporal graph dataset, enabling the model to convert text into structured graphs for more effective temporal reasoning. 
Additionally, \textit{TempReason}~\cite{yang-etal-2024-enhancing-temporal} leverages reinforcement learning to encourage the model to generate temporally coherent predictions, thereby improving the performance across diverse temporal reasoning tasks.
These efforts advance the temporal reasoning capabilities of LLMs across various tasks.
\section{SPAN: A Cross-Calendar Temporal Reasoning Benchmark for Large Language Models}
In this section, we first formalize the cross-calendar temporal reasoning task. Then, we describe the data acquisition and template design, including cross-calendar data collection, a description of the newly developed cross-calendar conversion interface, and the construction of question–code template pairs. Finally, we present a novel evaluation protocol for dynamic instance generation.
\begin{table*}[!t]
\setlength\tabcolsep{7pt}
	\renewcommand
	\arraystretch{1.4}
	\centering
    \begin{tabularx}{\textwidth}{@{}>{\centering\arraybackslash}p{0.15\textwidth}|>{\centering\arraybackslash}p{0.15\textwidth}|>{\raggedright\arraybackslash}X@{}}
    \cline{1-3} % 使用 booktabs 的横线，更美观
    \multicolumn{1}{c|}{\textbf{Reasoning Type}} & \multicolumn{1}{c|}{\textbf{Question Format}} & \multicolumn{1}{c}{\textbf{Question Template}} \\
    \cline{1-3}
    \multirow{16}{*}{\raisebox{-4pt}{\makecell{\strut Date- \\ based}}} & \multirow{8}{*}{\raisebox{-4pt}{\makecell{Content \\ question}}} & {\textit{Today's date on the $\{c_s\}$ calendar is ``$\{d^{r}_{c_{s}}\}$''. What was the $\{c_t\}$ calendar date $\{n_d\}$ days ago?}} \\
    \cline{3-3} % 只画第三列的横线
    & & {\textit{Today's date on the $\{c_s\}$ calendar is ``$\{d^{r}_{c_{s}}\}$''. What is the $\{c_t\}$ calendar date $\{n_d\}$ days later?}} \\
    \cline{3-3}
    & & {\textit{Today's date on the $\{c_s\}$ calendar is ``$\{d^{r}_{c_{s}}\}$''. What was the $\{c_t\}$ calendar date $\{n_w\}$ weeks ago?}} \\
    \cline{3-3}
    & & {\textit{Today's date on the $\{c_s\}$ calendar is ``$\{d^{r}_{c_{s}}\}$''. What is the $\{c_t\}$ calendar date $\{n_w\}$ weeks later?}} \\
    \cline{2-3}
    & \multirow{8}{*}{\makecell{Polar \\ question}} & {\textit{Today's date on the $\{c_s\}$ calendar is ``$\{d^{r}_{c_{s}}\}$''. Was the $\{c_t\}$ calendar date $\{n_d\}$ days ago equivalent to the date ``$\{d^e_{c_{t}}\}$''?}} \\
    \cline{3-3}
    & & {\textit{Today's date on the $\{c_s\}$ calendar is ``$\{d^{r}_{c_{s}}\}$''. Is the $\{c_t\}$ calendar date $\{n_d\}$ days later equivalent to the date ``$\{d^e_{c_{t}}\}$''?}} \\
    \cline{3-3}
    & & {\textit{Today's date on the $\{c_s\}$ calendar is ``$\{d^{r}_{c_{s}}\}$''. Was the $\{c_t\}$ calendar date $\{n_w\}$ weeks ago equivalent to the date ``$\{d^e_{c_{t}}\}$''?}} \\
    \cline{3-3}
    & & {\textit{Today's date on the $\{c_s\}$ calendar is ``$\{d^{r}_{c_{s}}\}$''. Is the $\{c_t\}$ calendar date $\{n_w\}$ weeks later equivalent to the date ``$\{d^e_{c_{t}}\}$''?}} \\
    \cline{1-3}
    \multirow{8}{*}{\makecell{Festival-\\based}} & \multirow{4}{*}{\makecell{Content \\ question}} & {\textit{Today's date on the $\{c_s\}$ calendar is ``$\{d^{r}_{c_{s}}\}$''. What was the $\{c_t\}$ calendar date of the $\{c_s\}$ festival ``$\{f_{c_s}\}$'' $\{n_d\}$ ago?}} \\
    \cline{3-3}
    & & {\textit{Today's date on the $\{c_s\}$ calendar is ``$\{d^{r}_{c_{s}}\}$''. What is the $\{c_t\}$ calendar date of the $\{c_s\}$ festival ``$\{f_{c_s}\}$'' $\{n_y\}$ later?}} \\
    \cline{2-3}
    & \multirow{4}{*}{\makecell{Polar \\ question}} & {\textit{Today's date on the $\{c_s\}$ calendar is ``$\{d^{r}_{c_{s}}\}$''. Was the $\{c_t\}$ calendar date of the $\{c_s\}$ festival ``$\{f_{c_s}\}$'' $\{n_y\}$ years ago equivalent to the date "$\{d^e_{c_{t}}\}$"?}} \\
    \cline{3-3}
    & & {\textit{Today's date on the {$\{c_s\}$} calendar is ``$\{d^{r}_{c_{s}}\}$''. Is the {$\{c_t\}$} calendar date of the $\{c_s\}$ festival ``$\{f_{c_s}\}$'' $\{n_y\}$ years later equivalent to the date ``$\{d^e_{c_{t}}\}$''?}} \\
    \cline{1-3}
    \end{tabularx}
    \vspace{4pt}
    \caption{
           The question templates. We use curly-brace placeholders to denote variables: $c_s$ and $c_t$ denote the source and target calendars; $f_{c_s}$ denotes the festival in the source calendar; $d^{r}_{c_{s}}$ denotes the reference date in the source calendar; $d^e_{c_{t}}$ denotes the expected date in the target calendar (used for polar questions); $n_d$, $n_w$, and $n_y$ denote temporal offsets in days, weeks, and years, respectively.
      }\label{table:question template}
\end{table*}
\begin{figure*}[!t]
    \centering % 将图片置于中央
    \includegraphics[width=0.85\linewidth]{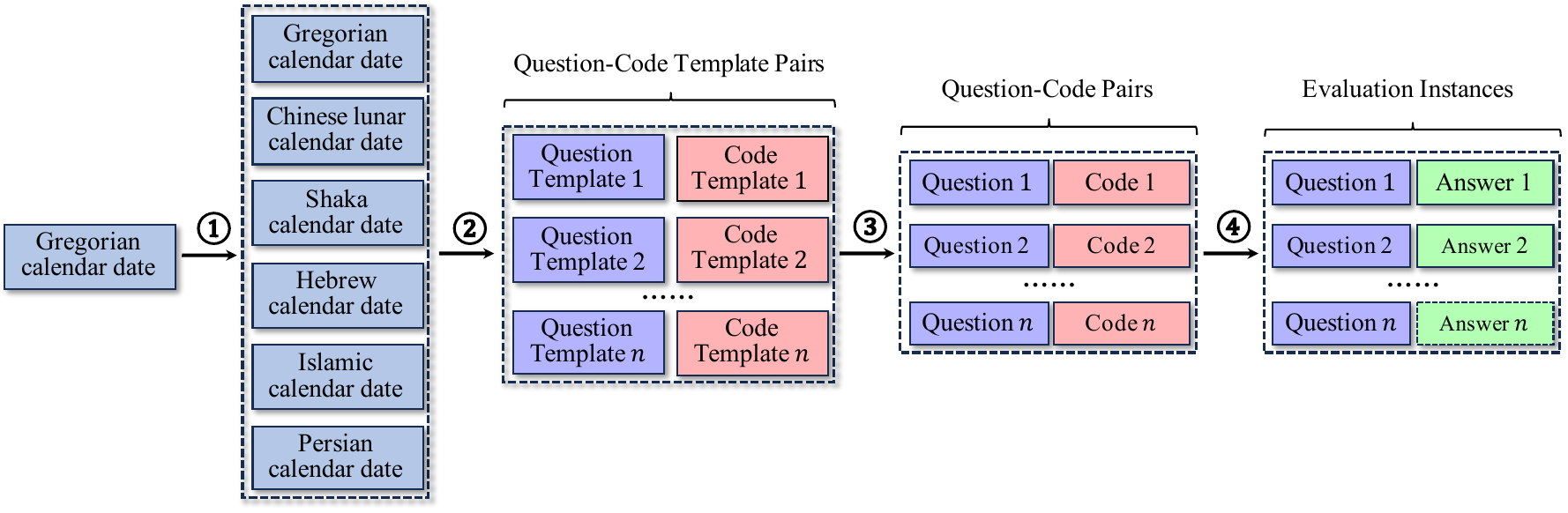}
    \caption{Overview of the proposed evaluation protocol. Given a user-specified Gregorian date as input, the process proceeds through four stages:~\ding{192}~\textbf{\textit{Calendar Conversion}}. The Gregorian date is converted into its equivalents in five calendars via our \texttt{search\_calendar} interface, yielding $(c_s, d_{c_s}^r, f_{c_s})$ pairs, with $d_{c_s}^r$ and $f_{c_s}$ denoting the reference date and the festival in the source calendar $c_s$, respectively.~\ding{193}~\textbf{\textit{Template Matching}}. These pairs are further utilized to construct $(c_s, d_{c_s}^r, f_{c_s}, c_t)$ pairs. Here, $c_s$ and $c_t$ are selected specifically to ensure one is a Gregorian calendar and the other is a non-Gregorian calendar. These pairs are matched against all question–code template pairs to generate candidate pairs.~\ding{194}~\textbf{\textit{Template Instantiation}}. For each candidate question-code template pair, we manually specify the remaining variables $(d_{c_t}^e, n_d, n_w, n_y)$. Afterwards, question–code pairs are generated by filling the template placeholders with all variables.~\ding{195}~\textbf{\textit{Code Execution}}. Finally, we execute each code snippet to generate the gold answer. }\label{fig:data_construction_pipeline}
  \end{figure*}
\subsection{Task Formulation}\label{subsection:problem_formulation}
Typically, the cross-calendar temporal reasoning task involves both intra-calendar date reasoning and inter-calendar date conversion, requiring LLMs to understand temporal relationships within individual calendars and to grasp temporal alignments between different calendars. Formally, this process can be represented as follows:
\begin{equation}
{d_{c_s}^{r}} 
\xrightarrow{\texttt{Reasoning}} 
{d_{c_s}}
\xrightarrow{\texttt{Converting}} 
{d_{c_t}^{t}} 
\tag{1}\label{eq:reasoning_chain}
\end{equation}
where ${d_{c_s}^{r}}$ and ${d_{c_s}}$ represent the reference date and reasoning date in the source calendar ${c_{s}}$, respectively; ${d_{c_t}^{t}}$ denotes the target date derived by converting ${d_{c_s}^{i}}$ to the target calendar ${c_{t}}$. Note that Equation~(\ref{eq:reasoning_chain}) is symmetric, enabling the source and target calendars to be interchanged. 

In this work, we focus on a practical and representative cross-calendar temporal reasoning scenario, namely, reasoning between the Gregorian calendar and other calendar systems, which is prevalent in many countries and regions~\cite{taqizadeh1939various,graumann2015problem}. Accordingly, in our formulation, when either the source calendar ${c_{s}}$ or the target calendar ${c_{t}}$ is set to the Gregorian calendar, the other is set to a non-Gregorian calendar.
\subsection{Data Collection and Template Design}\label{subsection:data_collection_and_curation}
\paragraph{Cross-Calendar Data Collection.} SPAN covers dates from three major calendar families: solar~(\textit{i.e.}, the Gregorian, Persian and Shaka calendars), lunar~(\textit{i.e.}, the Islamic calendar), and lunisolar~(\textit{i.e.}, the Hebrew and Chinese lunar calendars). Calendar dates are collected using Python libraries and \texttt{Wikipedia}\footnote{https://www.wikipedia.org}. Specifically, the \texttt{datetime} library\footnote{https://docs.python.org/3/library/datetime.html} is employed to enumerate Gregorian dates, and these dates are subsequently converted to their equivalent Islamic, Hebrew, Shaka, and Persian calendar dates using the \texttt{convertdate} library\footnote{https://pypi.org/project/convertdate}, while the conversions to the Chinese lunar calendar are handled using the \texttt{LunarCalendar} library\footnote{https://pypi.org/project/LunarCalendar}. Additionally, festival dates for each calendar are sourced and curated from \texttt{Wikipedia}. Festivals are listed in Appendix A. Finally, the collected dates are aligned to construct cross-calendar entries, each containing equivalent dates across the aforementioned six calendars.
\paragraph{Cross-Calendar Conversion Interface.} We develop a unified interface, \texttt{search\_calendar}, which converts a date or festival in the given calendar to its equivalents in multiple target calendars. The interface supports two parameter schemas: \{\texttt{calendar\_name}, \texttt{year}, \texttt{month}, \texttt{day}\} to specify a date, and \{\texttt{calendar\_name}, \texttt{year}, \texttt{festival\_name}\} to specify a festival in the given calendar. Upon invocation, $\texttt{search\_calendar}$ returns a cross-calendar entry containing equivalent dates across six calendars, enabling subsequent temporal computations. In this work, we employ this interface in both the code templates and the proposed \texttt{Time}~\texttt{Agent}. We provide a detailed description of \texttt{search\_calendar} in Appendix D.
\paragraph{Question Template Design.}
As shown in Table~\ref{table:question template}, we design a set of question templates, which are organized into two categories: 
(1) \textbf{\textit{Date-Based Question Templates}}, which apply temporal offsets in days or weeks to a reference date and convert the resulting date to its equivalent in the target calendar. (2) \textbf{\textit{Festival-Based Question Templates}}, which apply a year offset to a reference year, identify the corresponding festival date for the resulting year, and convert it to its equivalent in the target calendar. For each category, two question formats are included: polar questions (answered with ``Yes'' or ``No'') and content questions (answered with specific dates). The question templates collectively define eight variables, as detailed below. $c_s$ and $c_t$ denote the source and target calendars; $d_{c_{s}}^r$ denotes the reference date in the source calendar; $d_{c_{t}}^e$ denotes the expected date in the target calendar (used for polar questions); $f_{c_s}$ denotes the festival in the source calendar; $n_d$, $n_w$, and $n_y$ denote temporal offsets in days, weeks, and years, respectively. Among these, $(c_s$, $c_t$, $d_{c_{s}}^r$, $f_{c_s})$ are determined during evaluation, whereas $(d_{c_{t}}^e$, $n_d$, $n_w$, $n_y)$ are manually configurable.
\paragraph{Code Template Design.} For each question template, we design a corresponding code template in Python for answer generation. Code templates mirror the date-related variables defined in the corresponding question templates but differ in how these variables are represented. Question templates express variables in natural language, whereas code templates represent them as code snippets. For example, the natural language expression ``$\{n_d\}$ days later'' corresponds to the code snippet ``$\texttt{+ timedelta(days=}\{n_d\}\texttt{)}$''. We divide the code templates into two categories based on the reasoning direction: \textit{Gregorian-to-Others} and \textit{Others-to-Gregorian}, indicating conversions from the Gregorian calendar to other calendar systems and vice versa, respectively. For the $\textit{Gregorian-to-Others}$ category, we first use the \texttt{datetime} library to compute the reasoning date in the Gregorian calendar based on the given reference date, and then convert this result to the target calendar using our $\texttt{search\_calendar}$ interface. Conversely, for the $\textit{Others-to-Gregorian}$ category, we first convert the reference date to the Gregorian calendar using our \texttt{search\_calendar} interface, and then perform further computations with the \texttt{datetime} library.
\subsection{Evaluation Protocol for Dynamic Instance Generation}\label{subsection:evaluation_instance_generation_protocol}
Building upon our prior data preparation, we propose a novel evaluation protocol for on-the-fly instance generation, as illustrated in Figure~\ref{fig:data_construction_pipeline}. 
This protocol takes a user-specified Gregorian date as input and produces a diverse set of instances for cross-calendar temporal reasoning. Specifically, this protocol consists of the following stages: 
\ding{192}~\textbf{\textit{Calendar Conversion}}. We utilize the \texttt{search\_calendar} interface to convert the user-specific Gregorian date into its equivalents in the Chinese lunar, Islamic, Hebrew, Shaka, and Persian calendars. This yields a set of $(c_s, d_{c_s}^r,f_{c_s})$ pairs, where $d_{c_s}^r$ and $f_{c_s}$ denote the reference date and the festival in the source calendar $c_s$, respectively. \ding{193}~\textbf{\textit{Template Matching}}. Building on these pairs, we further construct $(c_s, d_{c_s}^r, f_{c_s}, c_t)$ pairs, where $c_s$ and $c_t$ are selected such that one is the Gregorian calendar and the other is a non-Gregorian calendar. These pairs are then matched with all question-code template pairs to yield candidate pairs for subsequent instantiation. \ding{194}~\textbf{\textit{Template Instantiation}}. For each question-code candidate template pair, we manually specify the remaining variables $(d_{c_{t}}^e, n_d, n_w, n_y)$ from a predefined set (see the \textit{Experimental Setup} for details). These variables are then substituted into the corresponding placeholders in the question and code templates, producing paired questions and executable code. \ding{195}~\textbf{\textit{Code Execution}}. Finally, each code snippet is executed, and its output serves as the ground truth answer for the corresponding evaluation instance.
\section{Experimental Setup}\label{section:experimental_setup}
\paragraph{Evaluation Models.} We evaluate six open- and closed-source LLMs. Specifically, the open-source models include \texttt{Llama-3.3-70B-Instruct}~\cite{grattafiori2024llama}, \texttt{DeepSeek-V3-1226}~\cite{liu2024deepseek}, and \texttt{Qwen-2.5-72B-Instruct}~\cite{yang2024qwen2}, which are accessed via Hugging Face and deployed using the vLLM framework~\cite{kwon2023efficient}; the closed-source models include \texttt{GPT-4o}~\cite{hurst2024gpt}, \texttt{Claude-3.7-Sonnet}~\cite{anth2025claude}, and \texttt{Gemini-1.5-Pro}~\cite{team2024gemini}, which are accessed through their APIs. All models are evaluated using greedy decoding with a temperature of $0.0$ to ensure reproducibility.
\paragraph{Evaluation Metric.} To enable reliable and scalable evaluation, we adopt GPT-4o as an automatic evaluator. Given a question, a model-generated response, and the gold answer, GPT-4o is prompted to assess the correctness of the response. Accuracy is then computed over the entire test set based on GPT-4o's judgments. The complete evaluation prompt is detailed in Appendix E, and we also present the agreement between the GPT-4o evaluator and human annotations in Appendix C.
\begin{figure}[!t]
  \centering
  \includegraphics[width=1.0\linewidth]{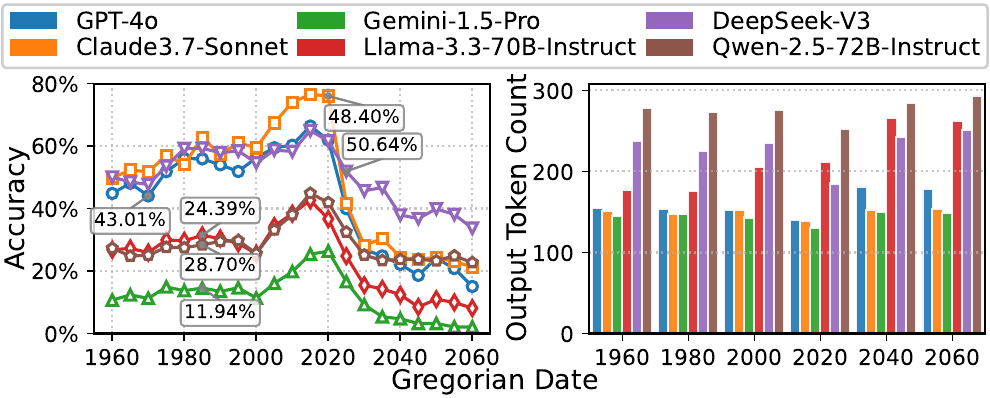}
  \caption{Left: Accuracy of LLMs across evaluation dates ranging from July $1$st, $1960$ to July $1$st, $2060$ at five-year intervals (July $1$st omitted for clarity). The average accuracy over time is annotated for each model. Right: Average output token counts of LLMs at sampled evaluation dates. To ensure comparability, model outputs are tokenized using OpenAI’s \texttt{tiktoken} tokenizer with the \texttt{o200k\_base} encoding.}\label{fig:main_result}
\end{figure}
\paragraph{Evaluation Setup.} The variables ($c_s$, $d_{c_s}^r$, $f_{c_s}$ $c_t$) are dynamically determined by our evaluation protocol, while the remaining variables ($d_{c_t}^e$, $n_d$, $n_w$, $n_y$)  are manually specified as follows. The offset variables $n_d$ and $n_w$ are set to integers in the range [$1$, $10$], and $n_y$ in the range [$1$, $5$]. For polar questions, the variable $d_{c_t}^e$ is set to the answer date, ensuring that all polar questions yield “Yes” as the gold answer. Additionally, we set the user-specified Gregorian dates~(evaluation dates) to July $1$st of every fifth year from $1960$ to $2060$, resulting in $21$ evaluation dates over $100$ years. At each of the $21$ evaluation dates, $1$,$780$ instances are generated, comprising $800$ date-based and $980$ festival-based questions, yielding a total of $37$,$380$ evaluation instances. Finally, all numeric month values in the generated instances are replaced with their corresponding textual month names according to their respective calendars.
\subsection{Main Results}\label{subsection:overall_results}The left part of Figure \ref{fig:main_result} shows accuracy at five-year intervals from July $1$st, $1960$ to July $1$st, $2060$. The following conclusions are drawn based on empirical evidence:\\
\textbf{\textit{Cross-Calendar Temporal Reasoning Remains Challenging.}}
As shown in the right part of Figure~\ref{fig:main_result}, we observe that LLMs tend to produce lengthy outputs ($150$$\sim$$300$ tokens), indicating that they perform explicit reasoning rather than directly predicting dates for cross-calendar temporal questions. Despite these reasoning attempts, the average accuracy across LLMs and evaluation dates remains only $34.5$\%, with none exceeding $80$\% accuracy across evaluation dates.
This underscores substantial room for improvement in LLMs' cross-calendar temporal reasoning capabilities.\\
\textbf{\textit{Accuracy Stratification in Large Language Models.}}
We find that the closed-source LLMs (\textit{e.g.}, \texttt{GPT-4o} and \texttt{Claude-3.7-Sonnet}) generally outperform the open-source ones (\textit{e.g.}, \texttt{Llama-3.3-70B-Instruct} and \texttt{Qwen-2.5-72B-Instruct}).
Encouragingly, the open-source \texttt{DeepSeek-V3} achieves accuracy on par with leading closed-source LLMs, whereas the closed-source \texttt{Gemini-1.5-Pro} demonstrates the poorest performance, with accuracy consistently below $30$\% across evaluation dates. 
These observations reflect the strengths of commercial LLMs and progress in the open-source community.\\
\textbf{\textit{Accuracy Exhibits Significant Temporal Dependency.}} 
Across all LLMs, accuracies tend to improve for past dates but drop sharply for future ones. 
This temporally-dependent pattern of accuracy also persists across different reasoning types, question formats, and temporal reasoning directions~(see the \textit{Analysis} section for details). These findings suggest that LLMs have a limited capability to perform cross-calendar temporal reasoning for future dates, a phenomenon we term \textbf{\textit{Future-Date Degradation}}.
\section{Experimental Results}
\begin{figure}[!t]
  \centering
  \includegraphics[width=1.0\linewidth]{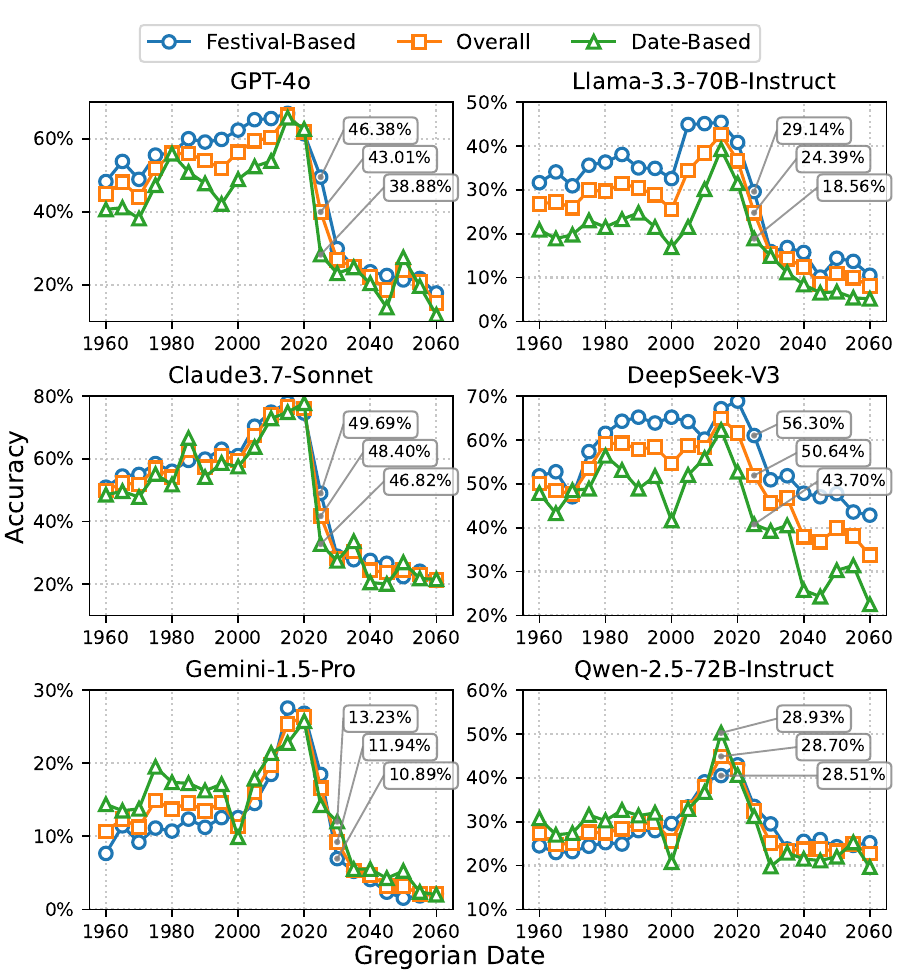}
  \caption{Accuracy of date-based and festival-based cross-calendar temporal reasoning over the evaluation dates from July $1$st, $1960$ to July $1$st, $2060$ at five-year intervals (July $1$st omitted for clarity). The average accuracy over time for each reasoning type is annotated.}\label{fig:effect_of_reasoning_type}
\end{figure}
\begin{figure}[!t]
  \centering
  \includegraphics[width=1.0\linewidth]{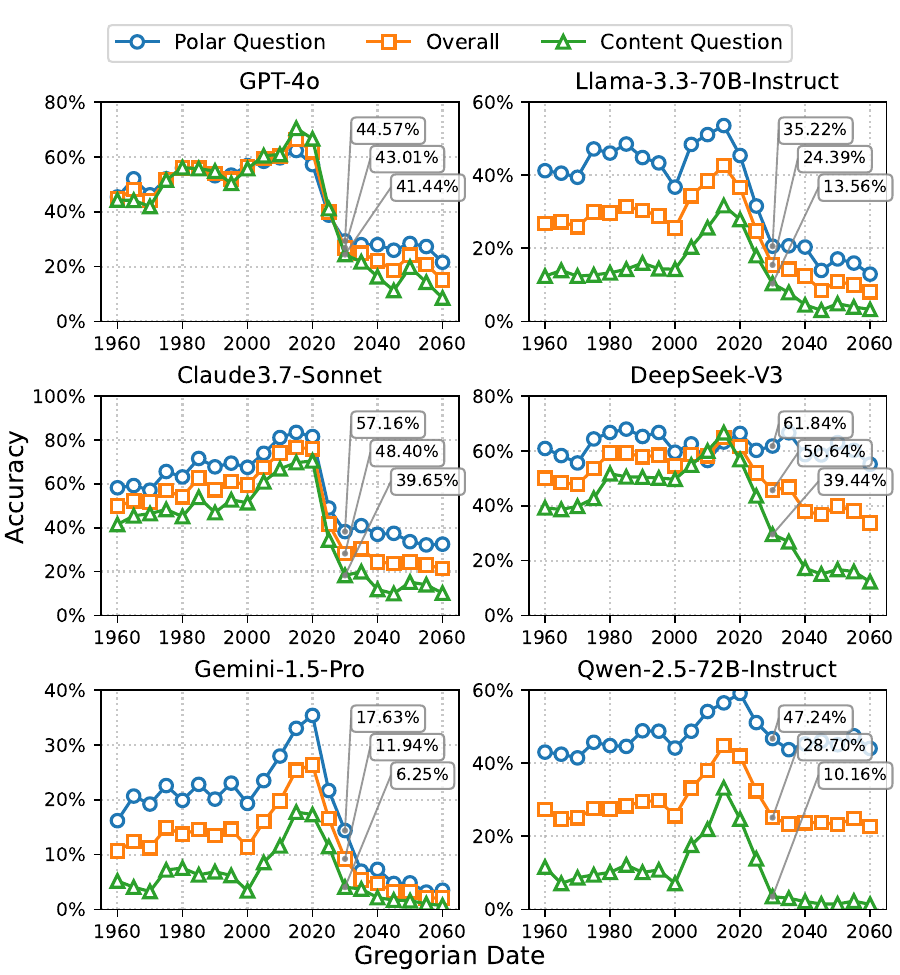}
  \caption{Accuracy of content question and polar question over the evaluation dates from July $1$st, $1960$ to July $1$st, $2060$ at five-year intervals (July $1$st omitted for clarity). The average accuracy over time for each question format is annotated.}\label{fig:effect_of_question_formats}
\end{figure}
\begin{figure}[!t]
  \centering
  \includegraphics[width=1.0\linewidth]{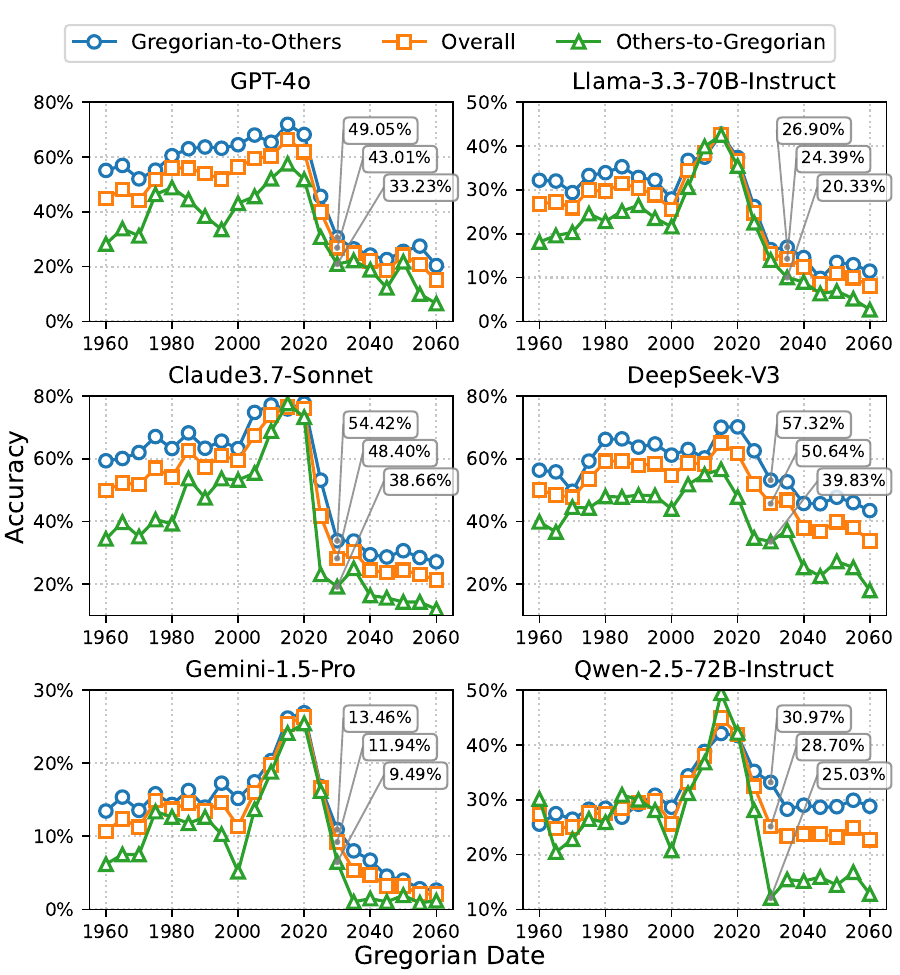}
  \caption{Accuracy of \textit{Gregorian-to-Others} and \textit{Others-to-Gregorian} cross-calendar temporal reasoning over the evaluation dates from July $1$st, $1960$ to July $1$st, $2060$ at five-year intervals (July $1$st omitted for clarity). The average accuracy over time for each group's temporal reasoning directions is annotated.}\label{fig:effect_of_calendar_direction}
\end{figure}
\subsection{Analysis}\label{subsection:analysis}
We empirically investigate the impact of various factors on LLMs' cross-calendar temporal reasoning capabilities, including reasoning types, question formats, and temporal reasoning directions. All comparisons reported have undergone significance testing, with detailed results provided in Appendix C.
\paragraph{\textbf{\textit{Date-Based vs. Festival-Based.}}} 
Figure~\ref{fig:effect_of_reasoning_type} presents the accuracy of LLMs on two reasoning types. We observe that most LLMs perform better on festival-based reasoning, with gains between $2.87$\% and $12.60$\%, except for \texttt{Gemini-1.5-Pro} and \texttt{Qwen-2.5-72B-Instruct}. This likely stems from the prevalence of festival dates in pretraining data, which reduces the difficulty of festival-based reasoning.
\paragraph{\textbf{\textit{Polar Question vs. Content Question.}}} Figure~\ref{fig:effect_of_question_formats} presents the accuracy of two question formats. We find that the accuracy on polar questions consistently exceeds that of content questions for all LLMs, with gains ranging from $3.13$\% to $37.08$\% ($18.86$\% on average). This gap reflects the lower complexity of polar questions, whose answers are binary, resulting in a $50$\% chance of correctness even without precise temporal reasoning.
\paragraph{\textbf{\textit{Gregorian-to-Others vs. Others-to-Gregorian.}}} Figure~\ref{fig:effect_of_calendar_direction} presents accuracy for two group of temporal reasoning directions: \textit{Gregorian-to-Others} and \textit{Others-to-Gregorian}. We see that all LLMs perform better on the \textit{Gregorian-to-Others} group, with accuracy gains ranging from $3.97$\% to $17.49$\%, particularly among higher-performing models like \texttt{DeepSeek-V3} ($17.49$\%), \texttt{GPT-4o} ($15.82$\%), and \texttt{Claude-3.7-Sonnet} ($15.76$\%). We refer to this discrepancy as \textbf{\textit{Calendar Asymmetry Bias}} in LLMs, conjecturing that it likely stems from the prevalence of Gregorian-origin expressions in pretraining data. Since modern web documents and textual resources predominantly use Gregorian timestamps as the primary temporal anchor, LLMs are more frequently exposed to conversions originating from the Gregorian calendar, while receiving limited exposure to the reverse direction.
\section{Improving Cross-Calendar Temporal Reasoning of LLMs}\label{section:improved_method} In this section, we present a practical method to improve LLM performance on SPAN. Our experiments indicate that LLMs face challenges with direct reasoning in this task. Consequently, we rely on an external tool for date conversion and defer enhancing LLMs' intrinsic reasoning to future work, emphasizing a pragmatic solution here.
\paragraph{\textbf{Our Method.}} 
We develop an LLM-powered \texttt{Time}~\texttt{Agent} that combines LLM code-generation capabilities with our cross-calendar conversion interface \texttt{search\_calendar}. In particular, it follows a three-step agentic workflow: \ding{192}~GPT-4o is guided with the description of \texttt{search\_calendar} in a few-shot prompting, enabling it to generate executable code snippets for solving the given question. \ding{193}~The generated code snippet is executed via a code interpreter, and we obtain the execution results. \ding{194}~The execution results are appended to the dialogue context as additional input, upon which GPT-4o produces the final answer. The complete prompts of \texttt{Time}~\texttt{Agent} are provided in Appendix F.
\paragraph{\textbf{Baselines.}} 
We compare our method with the following competitive baselines:
\begin{itemize}
    \item \texttt{Previous}~\texttt{Best}~\texttt{Results}: The best results across all LLMs for each evaluation date, illustrated in Figure~\ref{fig:main_result}.
    \item \texttt{OpenAI-o1}: A state-of-the-art closed-source reasoning model with high token consumption, requiring extensive reasoning chains to process questions.
    \item \texttt{GPT-4o}~\texttt{w/}~\texttt{RAG}: The GPT-4o model augmented with a retrieval-augmented generation (RAG) system, leveraging the Bing Search\footnote{\url{https://www.microsoft.com/en-us/bing}} for external retrieval.
    \item \texttt{GPT-4o}: The GPT-4o model, which is used to assess the effects of introducing RAG.
  \end{itemize}
\begin{figure}[!t]
  \centering
  \includegraphics[width=1.0\linewidth]{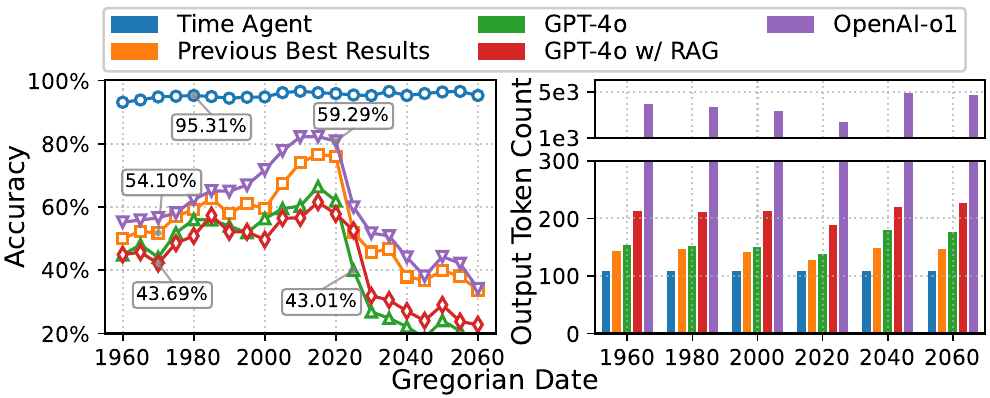}
  \caption{Left: Accuracy of each method over the evaluation dates at five-year intervals from July $1$st, $1960$ to July $1$st, $2060$. Right: Average output token count of each method at sampled evaluation dates. Model outputs are tokenized using the \texttt{tiktoken} library with the \texttt{o200k\_base} encoding.}\label{fig:our_method}
\end{figure}
\paragraph{\textbf{Results.}} As depicted in Figure~\ref{fig:our_method}, \texttt{Time}~\texttt{Agent} demonstrates consistently high accuracy across all evaluation dates, achieving an average accuracy of $95.31$\% while maintaining a minimal average output token count. 
The few errors in \texttt{Time}~\texttt{Agent} primarily stem from occasional code execution failures or subtle logical bugs, which can be further mitigated through enhanced prompt engineering. By comparison, \texttt{OpenAI-o1} attains the second-highest average accuracy of $59.29$\%, though it remains significantly below that of \texttt{Time}~\texttt{Agent} and incurs substantially greater token usage. Additionally, we observe that \texttt{GPT-4o}~\texttt{w/}~\texttt{RAG} achieves an average accuracy of $43.69$\%, and slightly outperforms \texttt{GPT-4o} on future evaluation dates, with an average improvement of merely $0.68$\%. However, this marginal gain requires generating more output tokens compared to \texttt{GPT-4o}, indicating that RAG provides negligible benefit for the cross-calendar temporal reasoning task. In summary, our results show that reasoning models and RAG are insufficient for the cross-calendar temporal reasoning task. By contrast, tool-augmented code generation provides a robust and efficient solution.
\section{Conclusion}
In this work, we introduce \textbf{SPAN}, a benchmark to evaluate the cross-calendar temporal reasoning capabilities of LLMs. Unlike prior benchmarks that are Gregorian-centric and time-invariant, SPAN enables cross-calendar temporal reasoning across ten reasoning directions, two reasoning types, and two question formats. To support evaluation at different temporal contexts, we design a novel template-driven protocol for time-variant evaluation. Based on this protocol, we perform extensive empirical evaluations on dates ranging from $1960$ to $2060$, revealing significant limitations in LLMs' capability to reason across calendars, with the challenges including \textit{Future-Date Degradation} and \textit{Calendar} \textit{Asymmetry} \textit{Bias}. 
Furthermore, we demonstrate that an LLM-powered \texttt{Time}~\texttt{Agent} with tool-augmented code generation achieves impressive performance on the cross-calendar temporal reasoning task.

In future work, we plan to extend SPAN to encompass a wider spectrum of calendar systems, enabling more comprehensive coverage of global temporal contexts. Moreover, we will adapt existing Gregorian-based temporal reasoning tasks (\textit{e.g.}, event ordering and duration estimation) to cross-calendar settings, facilitating a systematic evaluation of LLMs’ generalization capabilities and advancing their robustness in diverse temporal reasoning scenarios.
\section{Acknowledgments}
This work was independently conducted at Li Auto. We sincerely thank Li Auto for its generous support, which was essential to the successful completion of this work.
\bibliography{aaai2026}

@inproceedings{chen2021datasetansweringtimesensitivequestions,
  title={A Dataset for Answering Time-Sensitive Questions},
  author={Chen, Wenhu and Wang, Xinyi and Wang, William Yang},
  booktitle={Proceedings of NeurIPS},
  year={2021}
}

@inproceedings{tan2023benchmarkingimprovingtemporalreasoning,
      title={Towards Benchmarking and Improving the Temporal Reasoning Capability of Large Language Models}, 
      booktitle={Proceedings of ACL},
      author={Qingyu Tan and Hwee Tou Ng and Lidong Bing},
      year={2023}
}

@inproceedings{wang2024trambenchmarkingtemporalreasoning,
      title={TRAM: Benchmarking Temporal Reasoning for Large Language Models}, 
      booktitle={Proceedings of ACL (Findings)},
      author={Yuqing Wang and Yun Zhao},
      year={2024}
}

@article{graumann2015problem,
  title={The problem field of calendars in different cultures},
  author={Graumann, G{\"u}nter},
  journal={LUMAT: International Journal on Math, Science and Technology Education},
  year={2015}
}

@inproceedings{deng-etal-2024-investigating,
    title = "Investigating Data Contamination in Modern Benchmarks for Large Language Models",
    author = "Deng, Chunyuan  and
      Zhao, Yilun  and
      Tang, Xiangru  and
      Gerstein, Mark  and
      Cohan, Arman",
    booktitle = "Proceedings of ACL",
    year = "2024",
}

@inproceedings{sainz-etal-2023-nlp,
    title = "{NLP} Evaluation in trouble: On the Need to Measure {LLM} Data Contamination for each Benchmark",
    author = "Sainz, Oscar  and
      Campos, Jon  and
      Garc{\'i}a-Ferrero, Iker  and
      Etxaniz, Julen  and
      de Lacalle, Oier Lopez  and
      Agirre, Eneko",
    booktitle = "Proceedings of EMNLP (Findings)",
    year = "2023"
}

@inproceedings{balloccu2024leak,
  title={Leak, Cheat, Repeat: Data Contamination and Evaluation Malpractices in Closed-Source LLMs},
  author={Balloccu, Simone and Schmidtov{\'a}, Patr{\'\i}cia and Lango, Mateusz and Du{\v{s}}ek, Ond{\v{r}}ej},
  booktitle={Proceedings of ECAL},
  year={2024}
}

@inproceedings{sutimo,
  title={Timo: Towards Better Temporal Reasoning for Language Models},
  author={Su, Zhaochen and Zhang, Jun and Zhu, Tong and Qu, Xiaoye and Li, Juntao and Cheng, Yu and others},
  booktitle={Proceedings of COLM},
year={2024}
}

@article{taqizadeh1939various,
  title={Various Eras and Calendars used in the Countries of Islam.},
  journal={Bulletin of the School of Oriental and African Studies},
  year={1939},
  author={Taqizadeh, SH}
}

@inproceedings{li2024understanding,
  title={Understanding and Patching Compositional Reasoning in LLMs},
  author={Li, Zhaoyi and Jiang, Gangwei and Xie, Hong and Song, Linqi and Lian, Defu and Wei, Ying},
  booktitle={Proceedings of ACL (Findings)},
  year={2024}
}

@inproceedings{xu2024faithful,
  title={Faithful Logical Reasoning via Symbolic Chain-of-Thought},
  author={Xu, Jundong and Fei, Hao and Pan, Liangming and Liu, Qian and Lee, Mong-Li and Hsu, Wynne},
  booktitle={Proceedings of ACL},
  year={2024}
}

@inproceedings{fan-etal-2025-slam,
    title = "{SLAM}: Towards Efficient Multilingual Reasoning via Selective Language Alignment",
    author = "Fan, Yuchun  and
      Mu, Yongyu  and
      Wang, YiLin  and
      Huang, Lei  and
      Ruan, Junhao  and
      Li, Bei  and
      Xiao, Tong  and
      Huang, Shujian  and
      Feng, Xiaocheng  and
      Zhu, Jingbo",
    booktitle = "Proceedings of COLING",
    year = "2025",
}

@inproceedings{qiao-etal-2023-reasoning,
    title = "Reasoning with Language Model Prompting: A Survey",
    author = "Qiao, Shuofei  and
      Ou, Yixin  and
      Zhang, Ningyu  and
      Chen, Xiang  and
      Yao, Yunzhi  and
      Deng, Shumin  and
      Tan, Chuanqi  and
      Huang, Fei  and
      Chen, Huajun",
    booktitle = "Proceedings of ACL",
    year = "2023",
}

@article{saxena2025losttimeclockcalendar,
      title={Lost in Time: Clock and Calendar Understanding Challenges in Multimodal LLMs}, 
      author={Rohit Saxena and Aryo Pradipta Gema and Pasquale Minervini},
      year={2025},
      journal={arXiv preprint arXiv:2502.05092},
}

@inproceedings{jain2023language,
  title={Do language models have a common sense regarding time? revisiting temporal commonsense reasoning in the era of large language models},
  author={Jain, Raghav and Sojitra, Daivik and Acharya, Arkadeep and Saha, Sriparna and Jatowt, Adam and Dandapat, Sandipan},
  booktitle={Proceedings of EMNLP},
  year={2023}
}

@article{liu2024deepseek,
  title={Deepseek-v3 technical report},
  author={Liu, Aixin and Feng, Bei and Xue, Bing and Wang, Bingxuan and Wu, Bochao and Lu, Chengda and Zhao, Chenggang and Deng, Chengqi and Zhang, Chenyu and Ruan, Chong and others},
  journal={arXiv preprint arXiv:2412.19437},
  year={2024}
}

@article{grattafiori2024llama,
  title={The llama 3 herd of models},
  author={Grattafiori, Aaron and Dubey, Abhimanyu and Jauhri, Abhinav and Pandey, Abhinav and Kadian, Abhishek and Al-Dahle, Ahmad and Letman, Aiesha and Mathur, Akhil and Schelten, Alan and Vaughan, Alex and others},
  journal={arXiv preprint arXiv:2407.21783},
  year={2024}
}

@article{yang2024qwen2,
  title={Qwen2. 5 technical report},
  author={Yang, An and Yang, Baosong and Zhang, Beichen and Hui, Binyuan and Zheng, Bo and Yu, Bowen and Li, Chengyuan and Liu, Dayiheng and Huang, Fei and Wei, Haoran and others},
  journal={arXiv preprint arXiv:2412.15115},
  year={2024}
}

@article{hurst2024gpt,
  title={Gpt-4o system card},
  author={Hurst, Aaron and Lerer, Adam and Goucher, Adam P and Perelman, Adam and Ramesh, Aditya and Clark, Aidan and Ostrow, AJ and Welihinda, Akila and Hayes, Alan and Radford, Alec and others},
  journal={arXiv preprint arXiv:2410.21276},
  year={2024}
}

@article{anth2025claude,
  title={Claude 3.7 sonnet and claude code.},
  author={Anthropic},
  journal={https://www.anthropic.com/news/claude-3-7-sonnet},
  year={2025}
}

@article{team2024gemini,
  title={Gemini 1.5: Unlocking multimodal understanding across millions of tokens of context},
  author={Team, Gemini and Georgiev, Petko and Lei, Ving Ian and Burnell, Ryan and Bai, Libin and Gulati, Anmol and Tanzer, Garrett and Vincent, Damien and Pan, Zhufeng and Wang, Shibo and others},
  journal={arXiv preprint arXiv:2403.05530},
  year={2024}
}

@inproceedings{kwon2023efficient,
  title={Efficient memory management for large language model serving with pagedattention},
  author={Kwon, Woosuk and Li, Zhuohan and Zhuang, Siyuan and Sheng, Ying and Zheng, Lianmin and Yu, Cody Hao and Gonzalez, Joseph and Zhang, Hao and Stoica, Ion},
  booktitle={Proceedings of SOSP},
  year={2023}
}

@inproceedings{saxena2025lost,
  title={Lost in Time: Clock and Calendar Understanding Challenges in Multimodal LLMs},
  author={Saxena, Rohit and Gema, Aryo Pradipta and Minervini, Pasquale},
  booktitle = {Proceedings of ICLR},
  year={2025}
}

@inproceedings{fatemi2024test,
  title={Test of time: A benchmark for evaluating llms on temporal reasoning},
  author={Fatemi, Bahare and Kazemi, Mehran and Tsitsulin, Anton and Malkan, Karishma and Yim, Jinyeong and Palowitch, John and Seo, Sungyong and Halcrow, Jonathan and Perozzi, Bryan},
  booktitle={Proceedings of ICLR},
  year={2024}
}

@inproceedings{wei-etal-2023-menatqa,
    title = "{M}enat{QA}: A New Dataset for Testing the Temporal Comprehension and Reasoning Abilities of Large Language Models",
    author = "Wei, Yifan  and
      Su, Yisong  and
      Ma, Huanhuan  and
      Yu, Xiaoyan  and
      Lei, Fangyu  and
      Zhang, Yuanzhe  and
      Zhao, Jun  and
      Liu, Kang",
    booktitle = "Proceedings of EMNLP (Findings)",
    year = "2023"
 }

@inproceedings{qin-etal-2021-timedial,
    title = "{TIMEDIAL}: Temporal Commonsense Reasoning in Dialog",
    author = "Qin, Lianhui  and
      Gupta, Aditya  and
      Upadhyay, Shyam  and
      He, Luheng  and
      Choi, Yejin  and
      Faruqui, Manaal",
    booktitle = "Proceedings of ACL",
    year = "2021"
 }

@inproceedings{tan-etal-2023-towards,
    title = "Towards Benchmarking and Improving the Temporal Reasoning Capability of Large Language Models",
    author = "Tan, Qingyu  and
      Ng, Hwee Tou  and
      Bing, Lidong",
    booktitle = "Proceedings of ACL",
    year = "2023"
}

@inproceedings{chu-etal-2024-timebench,
    title = "{T}ime{B}ench: A Comprehensive Evaluation of Temporal Reasoning Abilities in Large Language Models",
    author = "Chu, Zheng  and
      Chen, Jingchang  and
      Chen, Qianglong  and
      Yu, Weijiang  and
      Wang, Haotian  and
      Liu, Ming  and
      Qin, Bing",
    booktitle = "Proceedings of ACL",
    year = "2024",
}

@article{wei2025time,
  title={TIME: A Multi-level Benchmark for Temporal Reasoning of LLMs in Real-World Scenarios},
  author={Wei, Shaohang and Li, Wei and Song, Feifan and Luo, Wen and Zhuang, Tianyi and Tan, Haochen and Guo, Zhijiang and Wang, Houfeng},
  journal={arXiv preprint arXiv:2505.12891},
  year={2025}
}

@inproceedings{yang-etal-2024-enhancing-temporal,
    title = "Enhancing Temporal Sensitivity and Reasoning for Time-Sensitive Question Answering",
    author = {Yang, Wanqi  and Li, Yanda  and Fang, Meng  and Chen, Ling},
    booktitle = "Proceedings of EMNLP (Findings)",
    year = "2024",
}

@article{ge2025tremu,
  title={TReMu: Towards Neuro-Symbolic Temporal Reasoning for LLM-Agents with Memory in Multi-Session Dialogues},
  author={Ge, Yubin and Romeo, Salvatore and Cai, Jason and Shu, Raphael and Sunkara, Monica and Benajiba, Yassine and Zhang, Yi},
  journal={arXiv preprint arXiv:2502.01630},
  year={2025}
}

@inproceedings{wang-zhao-2024-tram,
    title = "{TRAM}: Benchmarking Temporal Reasoning for Large Language Models",
    author = "Wang, Yuqing  and
      Zhao, Yun",
    booktitle = "Proceedings of ACL (Findings)",
    year = "2024"
}

@inproceedings{liska2022streamingqa,
  title={Streamingqa: A benchmark for adaptation to new knowledge over time in question answering models},
  author={Liska, Adam and Kocisky, Tomas and Gribovskaya, Elena and Terzi, Tayfun and Sezener, Eren and Agrawal, Devang and D’Autume, Cyprien De Masson and Scholtes, Tim and Zaheer, Manzil and Young, Susannah and others},
  booktitle="Proceedings of ICML",
  year={2022}
}

@inproceedings{kasai2023realtime,
  title={REALTIME QA: what's the answer right now?},
  author={Kasai, Jungo and Sakaguchi, Keisuke and Takahashi, Yoichi and Le Bras, Ronan and Asai, Akari and Yu, Xinyan Velocity and Radev, Dragomir and Smith, Noah A and Choi, Yejin and Inui, Kentaro},
  booktitle={Proceedings of NeurIPS},
  year={2023}
}

@inproceedings{yang-etal-2023-upon,
    title = "Once Upon a Time in Graph: Relative-Time Pretraining for Complex Temporal Reasoning",
    author = "Yang, Sen  and
      Li, Xin  and
      Bing, Lidong  and
      Lam, Wai",
    booktitle = "Proceedings of EMNLP",
    year = "2023",
}

@inproceedings{xiong2024large,
  title={Large Language Models Can Learn Temporal Reasoning},
  author={Xiong, Siheng and Payani, Ali and Kompella, Ramana and Fekri, Faramarz},
  booktitle={Proceedings of ACL},
  year={2024}
}

\clearpage
\section{Appendix}
\subsection{A. Festival List}
Table~\ref{tab:festivals} lists a total of $21$ festivals from four calendar systems: Gregorian, Chinese lunar, Islamic, and Persian.
\begin{table}[!ht]
\centering
\small
\begin{tabular}{|c|l|}
\hline
\makecell{Calendar} & \makecell{Festival} \\
\hline
\makecell{Gregorian\\calendar}
 & \makecell[l]{- \texttt{Halloween} \\
- \texttt{Christmas Day} \\
- \texttt{New Year's Day} \\
- \texttt{Valentine's Day} \\
- \texttt{International Women's Day} \\
- \texttt{International Workers' Day} \\
- \texttt{International Children's Day} \\
} \\
\hline
\makecell{Chinese\\lunar calendar}
 & \makecell[l]{- \texttt{Chinese New Year} \\
- \texttt{Lantern Festival} \\
- \texttt{Dragon Boat Festival} \\
- \texttt{Chinese Valentine's Day} \\
- \texttt{Ghost Festival} \\
- \texttt{Mid-Autumn Festival}
} \\
\hline
\makecell{Islamic\\calendar}
 & \makecell[l]{- \texttt{Hijri New Year} \\
- \texttt{Isra and Mi'raj} \\
- \texttt{Eid al-Fitr} \\
- \texttt{Eid al-Adha}
} \\
\hline
\makecell{Persian\\calendar}
 & \makecell[l]{- \texttt{Persian New Year} \\
- \texttt{Sizdah Be-dar} \\
- \texttt{Mehregan Festival} \\
- \texttt{Tirgan Festival}
} \\
\hline
\end{tabular}
\vspace{0.5ex}
\caption{Festivals in four calendars.}
\label{tab:festivals}
\end{table}
\subsection{B. Reliability of the GPT-4o Evaluator}
To validate the reliability of the GPT-4o evaluator, we conduct a correlation test. Concretely, we sample $1,260$ examples from different LLMs and compare GPT-4o's evaluations with human annotations using multiple correlation metrics.

As presented in Table~\ref{tab:agreement}, GPT-4o's evaluations exhibit a high degree of concordance with human annotations, achieving strong metric scores ranging from $0.95$ to $0.98$, with consistently high agreement across individual LLM outputs. These results support GPT-4o's reliability as an unbiased evaluator within our benchmark.
\begin{table}[h]
\centering
\setlength{\tabcolsep}{4pt}
\resizebox{\columnwidth}{!}{
\begin{tabular}{lcccc}
\toprule
\makecell{\textbf{Model}} & \makecell{\textbf{Agreement}\\ \textbf{Accuracy}} & \textbf{Spearman} & \textbf{Kendall} & \textbf{Cohen’s $\boldsymbol{\kappa}$}\\
\midrule
\makecell{-} & \makecell{$0.98$ \\ ($1232$/$1260$)} & $0.95$ & $0.95$ & $0.95$\\
\makecell{GPT-4o} & \makecell{$0.98$ \\ ($205$/$210$)} & $0.95$ & $0.95$ & $0.95$ \\
\makecell{Claude-3.7-\\Sonnet} & \makecell{$0.97$ \\ ($203$/$210$)} & $0.94$ & $0.94$ & $0.93$ \\
\makecell{Gemini-1.5-\\Pro} & \makecell{$0.97$ \\ ($204$/$210$)} & $0.88$ & $0.88$ & $0.87$ \\
\makecell{Llama-3.3-\\70B-Instruct} & \makecell{$0.96$ \\ ($201$/$210$)} & $0.88$ & $0.88$ & $0.88$ \\
\makecell{DeepSeek-V3} & \makecell{$1.00$ \\ ($210$/$210$)} & $1.00$ & $1.00$ & $1.00$ \\
\makecell{Qwen-2.5-\\72B-Instruct} & \makecell{$0.99$ \\ ($209$/$210$)} & $0.99$ & $0.99$ & $0.99$ \\
\bottomrule
\end{tabular}
}
\caption{Correlation metrics between the GPT-4o's evaluations and human annotations.}
\label{tab:agreement}
\end{table}
\subsection{C. Statistical Significance Testing}
We conduct significance testing using $10,000$ bootstrap iterations to rigorously evaluate the robustness of our experimental findings. Tables~\ref{tab:significance_festival}, \ref{tab:significance_polar}, and \ref{tab:significance_gregorian} present the results for the \textit{Festival-Based vs. Date-Based}, \textit{Polar-Question vs. Content-Question}, and \textit{Gregorian-to-Others vs. Others-to-Gregorian} reasoning tasks, respectively. All pairwise comparisons reveal statistically significant differences with $p$ $<$ $0.001$.
\begin{table}[h]
\centering
\setlength{\tabcolsep}{4pt}
\resizebox{\columnwidth}{!}{
\begin{tabular}{lccc}
\toprule
\makecell{\textbf{Model}} & \makecell{\textbf{Mean} \\ \textbf{Difference}} & \makecell{\textbf{95\%}\textbf{Confidence}\\ \textbf{Interval}} & \textbf{P-Value} \\
\midrule
\makecell{GPT-4o} & $0.16$ & [$0.15$, $0.17$] & $p$ $<$ $0.001$ \\
\makecell{Claude-3.7-Sonnet} & $0.16$ & [$0.15$, $0.17$] & $p$ $<$ $0.001$ \\
\makecell{Gemini-1.5-Pro} & $0.04$ & [$0.03$, $0.05$] & $p$ $<$ $0.001$ \\
\makecell{Llama-3.3-70B-Instruct} & $0.07$ & [$0.06$, $0.07$] & $p$ $<$ $0.001$ \\
\makecell{Deepseek-V3} & $0.17$ & [$0.16$, $0.19$] & $p$ $<$ $0.001$ \\
\makecell{Qwen-2.5-72B-Instruct} & $0.06$ & [$0.05$, $0.07$] & $p$ $<$ $0.001$ \\
\bottomrule
\end{tabular}
}
\caption{Significance test results ($10,000$ bootstrap iterations) for \textit{Festival-Based} and \textit{Date-Based} reasoning tasks.}
\label{tab:significance_festival}
\end{table}

\begin{table}[h]
\centering
\setlength{\tabcolsep}{4pt}
\resizebox{\columnwidth}{!}{
\begin{tabular}{lccc}
\toprule
\makecell{\textbf{Model}} & \makecell{\textbf{Mean} \\ \textbf{Difference}} & \makecell{\textbf{95\%}\textbf{Confidence}\\ \textbf{Interval}} & \textbf{P-Value} \\
\midrule
\makecell{GPT-4o} & $0.03$ & [$0.02$, $0.04$] & $p$ $<$ $0.001$ \\
\makecell{Claude-3.7-Sonnet} & $0.18$ & [$0.17$, $0.18$] & $p$ $<$ $0.001$ \\
\makecell{Gemini-1.5-Pro} & $0.11$ & [$0.11$, $0.12$] & $p$ $<$ $0.001$ \\
\makecell{Llama-3.3-70B-Instruct} & $0.22$ & [$0.21$, $0.22$] & $p$ $<$ $0.001$ \\
\makecell{Deepseek-V3} & $0.22$ & [$0.21$, $0.23$] & $p$ $<$ $0.001$ \\
\makecell{Qwen-2.5-72B-Instruct} & $0.37$ & [$0.36$, $0.38$] & $p$ $<$ $0.001$ \\
\bottomrule
\end{tabular}
}
\caption{Significance test results ($10,000$ bootstrap iterations) for \textit{Polar-Question} and \textit{Content-Question} reasoning tasks.}
\label{tab:significance_polar}
\end{table}

\begin{table}[h]
\centering
\setlength{\tabcolsep}{4pt}
\resizebox{\columnwidth}{!}{
\begin{tabular}{lccc}
\toprule
\makecell{\textbf{Model}} & \makecell{\textbf{Mean} \\ \textbf{Difference}} & \makecell{\textbf{95\%}\textbf{Confidence}\\ \textbf{Interval}} & \textbf{P-Value} \\
\midrule
\makecell{GPT-4o} & $0.16$ & [$0.02$, $0.04$] & $p$ $<$ $0.001$ \\
\makecell{Claude-3.7-Sonnet} & $0.16$ & [$0.15$, $0.17$] & $p$ $<$ $0.001$ \\
\makecell{Gemini-1.5-Pro} & $0.04$ & [$0.03$, $0.05$] & $p$ $<$ $0.001$ \\
\makecell{Llama-3.3-70B-Instruct} & $0.07$ & [$0.06$, $0.07$] & $p$ $<$ $0.001$ \\
\makecell{Deepseek-V3} & $0.17$ & [$0.16$, $0.19$] & $p$ $<$ $0.001$ \\
\makecell{Qwen-2.5-72B-Instruct} & $0.06$ & [$0.05$, $0.07$] & $p$ $<$ $0.001$ \\
\bottomrule
\end{tabular}
}
\caption{Significance test results ($10,000$ bootstrap iterations) for \textit{Gregorian-to-Others} and \textit{Others-to-Gregorian} reasoning tasks.}
\label{tab:significance_gregorian}
\end{table}

\subsection{D. Description of \textit{search\_calendar} Interface}
\begin{lstlisting}[language=json]
Description
--------
An interface for cross-calendar date 
conversion. Given a calendar and either a 
specific date or a festival, it maps the 
given date to equivalent dates in multiple 
target calendar systems.

Signature
----------
search_calendar(
    calendar_name: str,
    year: int,
    month: Optional[int] = None,
    day: Optional[int] = None,
    festival_name: Optional[str] = None,
    is_lunar_leap_month: bool = False
) -> dict[str, str]

Parameters
----------
- calendar_name : str
The source calendar (e.g., Gregorian)

- year: int
Year in the source calendar. (e.g., 1998)

- month: int, optional
Month in the source calendar. (e.g., 12)
Required when `day` is provided.

- day: int, optional
Day in the source calendar. (e.g., 23)
Required when `month` is provided.

- festival_name: str, optional
Name of a festival. Required when neither `
month` nor `day` is provided.

- is_lunar_leap_month: bool, optional (
default False)
Whether the given lunar month is a leap 
month. Ignored for calendars that do not 
distinguish leap months.

The interface supports two kinds of input:

- {calendar_name, year, month, day[, 
is_lunar_leap_month]}: Specify a concrete 
date in the source calendar.

- {calendar_name, year, festival_name}: 
Specify a festival in the source calendar.

The parameter `is_lunar_leap_month` is 
relevant for lunisolar calendars (e.g., 
Chinese Lunar calendar) where months can be 
leap months. It will be ignored for 
calendars that do not support this feature.

Return
-------
- a cross-calendar entry: dict
A dictionary mapping target calendars to 
their equivalent dates or festival dates for 
the given input.

Examples
--------
>>> search_calendar("Gregorian", 2024, 6, 1)
{'Gregorian': '2024-06-01', 'Chinese Lunar': 
'2024-04-25', ...}

>>> search_calendar("Chinese Lunar", 2025, 
festival_name="Chinese New Year")
{'Chinese Lunar': '2025-01-29', 'Gregorian': 
'2025-01-29', ...}
\end{lstlisting}

\subsection{E. Evaluation Prompt}
\begin{lstlisting}[language=json]
# Role: Text Evaluation Specialist

## Description
You are a Text Evaluation Specialist, your 
role is to assess response accuracy 
objectively based on task guidelines.

## Task

### Objective
Check if the `response` correctly answers 
the `question` using the `answer`.

### Inputs
Receive a tuple: (`question`, `response`, 
`answer`)
- `question`: A date conversion query.
- `response`: The given answer.
- `answer`: The correct answer.

### Evaluation Criteria
- **Accuracy**: Verify if `response` matches 
`answer`.
- **Ignore Formatting**: Disregard date 
format issues.

## Task Output
Use this JSON format for your evaluation, 1 
for correct, 0 for incorrect:
```json
{
  "accuracy": ""
}
```
\end{lstlisting}
\subsection{F. Prompts for Our Proposed Time Agent}
\noindent\textbf{1. Prompt for Code Generation}
\begin{lstlisting}[language=json]
# Role

## Calendar System Date and Festival 
Converter
You can accurately convert dates and 
festival observances across multiple 
calendar systems using the `search_calendar` 
function and functions from the `datetime` 
package.

## Function Definition

### Function: search_calendar

### Description
This function maps dates or festivals from a 
given calendar to their equivalents in 
multiple target calendars.

### Input Parameters
- `calendar`: which calendar to use 
[
    "gregorian", 
    "lunar", 
    "islamic", 
    "hebrew", 
    "shaka", 
    "persian"
]
- `year`: the year number
- `month`: the month number
- `day`: the day number
- `festival`: festival name
- `is_lunar_leap_month`: set True if `month` 
is a lunar leap month

### Usage Requirements
Provide either:
1. The full date (year, month, and day), or
2. The year and a valid festival name.

### Return
A dictionary with the date in multiple 
calendars. Example:
{
    "islamic_date": "1369-3-13",
    "lunar_date": "1949-11-15",
    "hebrew_date": "5710-10-14",
    "shaka_date": "1871-10-13",
    "persian_date": "1328-10-13",
    "gregorian_date": "1950-1-3"
}

## Supported Festivals

### Gregorian Calendar
[
    "New Year's Day", 
    "Valentine's Day", 
    "International Women's Day", 
    "International Workers' Day", 
    "International Children's Day", 
    "Halloween", "Christmas Day"
]

### Chinese Lunar Calendar
[
    "Chinese New Year", 
    "Lantern Festival", 
    "Dragon Boat Festival", 
    "Chinese Valentine's Day", 
    "Ghost Festival", 
    "Mid-Autumn Festival"
]

### Islamic Calendar
[
    "Hijri New Year", 
    "Isra and Mi'raj", 
    "Eid al-Fitr", 
    "Eid al-Adha"
]

### Persian Calendar
[
    "Persian New Year", 
    "Sizdah Be-dar", 
    "Mehregan Festival", 
    "Tirgan Festival"
]

## Month Name Mapping

### Hebrew Calendar
{
    1: "Nisan", 
    2: "Iyyar", 
    3: "Sivan", 
    4: "Tammuz", 
    5: "Av", 
    6: "Elul", 
    7: "Tishri", 
    8: "Heshvan", 
    9: "Kislev", 
    10: "Teveth", 
    11: "Shevat", 
    12: "Adar (Adar I)", 
    13: "Veadar (Adar II)"
}

### Shaka Calendar
{
    1: "Chaitra", 
    2: "Vaishakha", 
    3: "Jyeshtha", 
    4: "Ashadha", 
    5: "Shravana", 
    6: "Bhadrapada", 
    7: "Ashwin", 
    8: "Kartika", 
    9: "Margashirsha", 
    10: "Pausha", 
    11: "Magha", 
    12: "Phalguna"
}

### Islamic Calendar
{
    1: "Muharram", 
    2: "Safar", 
    3: "Rabi' al-Awwal", 
    4: "Rabi' al-Thani (Rabi' al-Akhir)", 
    5: "Jumada al-Awwal", 
    6: "Jumada al-Thani (Jumada al-Akhir)", 
    7: "Rajab", 
    8: "Sha'ban", 
    9: "Ramadan", 10: "Shawwal", 
    11: "Dhu al-Qi'dah", 
    12: "Dhu al-Hijjah"
}

### Persian Calendar
{
    1: "Farvardin", 
    2: "Ordibehesht", 
    3: "Khordad", 
    4: "Tir", 
    5: "Mordad", 
    6: "Shahrivar", 
    7: "Mehr", 
    8: "Aban", 
    9: "Azar", 
    10: "Dey", 
    11: "Bahman", 
    12: "Esfand"
}

# Workflow Guidance

## You can convert from Gregorian calendar 
to other calendars
1. Use the datetime package to process or 
calculate the Gregorian date.
2. Use search_calendar with the Gregorian 
result (calendar="gregorian").
3. Store the final answer in the variable 
answer on the last line. For date comparison
, refer to Example 3 or 4.

## You can convert from non-Gregorian
calendars to Gregorian calendar
1. Use search_calendar with the relevant 
non-Gregorian calendar and parameters.
2. If you need further date calculations, 
extract "gregorian_date" and use the 
datetime package.
3. Store the final answer in the variable 
answer on the last line. For date comparison
, refer to Example 3 or 4.

# Examples

## Example1
- Query: Today's date on the Gregorian 
calendar is "2020-7-1". What was the Islamic 
calendar date 7 days ago?

- Code: ```python
from datetime import datetime, timedelta

gregorian_date_today = datetime(2020, 7, 1)
gregorian_date_7_days_ago = \
gregorian_date_today - timedelta(days=7)

calendar = search_calendar(
    calendar="gregorian",
    year=gregorian_date_7_days_ago.year,
    month=gregorian_date_7_days_ago.month,
    day=gregorian_date_7_days_ago.day
)

answer = calendar.get("islamic_date")
```

## Example2
- Query: Today's date on the Chinese lunar 
calendar is "2020-7-1". What is the 
Gregorian calendar date of the Chinese lunar 
event "Mid-Autumn festival" 5 years later?

- Code: ```python
lunar_year = 2020 + 5

calendar = search_calendar(
    calendar="lunar", 
    year=lunar_year, 
    event="Mid-Autumn Festival"
)

answer = calendar.get("gregorian_date")
```

## Example3
- Query: Today's date on the Gregorian 
calendar is "1960-7-1". Was the Islamic 
calendar date 5 days ago equivalent to the 
date "1 Muharram 1380"?

- Code: ```python
from datetime import datetime, timedelta

gregorian_date_today = datetime(1960, 7, 1)
gregorian_date_5_days_ago = \
gregorian_date_today - timedelta(days=5)

calendar = search_calendar(
    calendar="gregorian",
    year=gregorian_date_5_days_ago.year,
    month=gregorian_date_5_days_ago.month,
    day=gregorian_date_5_days_ago.day
)

year, month, day = map(int, calendar. \
get("islamic_date").split("-"))

answer = bool(
    year==1380 and \
    month == 1 and \
    day == 1
)
```

## Example4
- Query: Today's date on the Hebrew calendar 
is "1 Tammuz 5725". Was the Gregorian 
calendar date 3 days ago equivalent to the 
date "1965-6-28"?

- Code: ```python
from datetime import datetime, timedelta
calendar = search_calendar(
    calendar="hebrew", 
    year=5725, 
    month=4, 
    day=1
)

gregorian_date_today = datetime.strptime( \
calendar.get("gregorian_date"), "%Y-%m-%d")

gregorian_date_3_days_ago = \
gregorian_date_today - timedelta(days=3)

answer = bool(
gregorian_date_3_days_ago.year == 1965 and \
gregorian_date_3_days_ago.month == 6 and \
gregorian_date_3_days_ago.day == 28")
```

# Output Restriction
Only respond with a code block enclosed in 
```python\n{code}\n``` and do not include 
any explanations or comments.
\end{lstlisting}
\noindent\textbf{2. Prompt for Final Response Generation}
\begin{lstlisting}[language=json]
Given a question and its answer, where the 
answer is either a numeric date string or a 
boolean, generate a response as follows:

1. Identify the calendar mentioned in the 
question (e.g., Gregorian, Hebrew, Shaka, 
Islamic, or Persian).

2. If the calendar is Hebrew, Shaka, Islamic
, or Persian, convert the numeric month in 
the answer to its corresponding month name 
using the mappings below and format the date 
as "{Month Name} {Day}, {Year}". For other 
calendars, keep the numeric date format.

3. If the answer is a boolean (True or False
), return "Yes." or "No." respectively.

Supported Calendars and Month Name Mappings:

Hebrew Calendar:
{
    1: "Nisan", 
    2: "Iyyar", 
    3: "Sivan", 
    4: "Tammuz", 
    5: "Av", 
    6: "Elul", 
    7: "Tishri", 
    8: "Heshvan", 
    9: "Kislev", 
    10: "Teveth", 
    11: "Shevat", 
    12: "Adar (Adar I)", 
    13: "Veadar (Adar II)"
}

Shaka Calendar:
{
    1: "Chaitra", 
    2: "Vaishakha", 
    3: "Jyeshtha", 
    4: "Ashadha", 
    5: "Shravana", 
    6: "Bhadrapada", 
    7: "Ashwin", 
    8: "Kartika", 
    9: "Margashirsha", 
    10: "Pausha", 
    11: "Magha", 
    12: "Phalguna"
}

Islamic Calendar:
{
    1: "Muharram", 
    2: "Safar", 
    3: "Rabi' al-Awwal", 
    4: "Rabi' al-Thani (Rabi' al-Akhir)", 
    5: "Jumada al-Awwal", 
    6: "Jumada al-Thani (Jumada al-Akhir)", 
    7: "Rajab", 8: "Sha'ban", 
    9: "Ramadan", 
    10: "Shawwal", 
    11: "Dhu al-Qi'dah", 
    12: "Dhu al-Hijjah"
}

Persian Calendar:
{
    1: "Farvardin", 
    2: "Ordibehesht", 
    3: "Khordad", 
    4: "Tir", 
    5: "Mordad", 
    6: "Shahrivar", 
    7: "Mehr", 
    8: "Aban", 
    9: "Azar", 
    10: "Dey", 
    11: "Bahman", 
    12: "Esfand"
}

Example1:
- Query: Today's date on the Gregorian 
calendar is "2060-7-1". What was the Islamic 
calendar date 10 days ago?
- Answer: 1483-1-22
- Response: 22 Muharram, 1483

Examples
Example2:
- Query: Today's date on the Islamic 
calendar is "2 Safar 1483". What was the 
Gregorian calendar date 1 days ago?
- Answer: 2060-06-30
- Response: 6 30, 2060

Output Requirement:
Provide only the final converted answer as a 
single English date string or "Yes"/"No" 
without additional explanations.
\end{lstlisting}

% \begin{tcolorbox}[breakable, enhanced, colback=gray!5, colframe=black!75, title=Prompt: Calendar Conversion Agent, fonttitle=\bfseries, listing only, listing options={basicstyle=\ttfamily\footnotesize, breaklines=true}]

% \subsection{Additional Experiments}

% Example: Insert a figure
% Uncomment and modify the following lines to add your own figures:
% \begin{figure}[h]
% \centering
% \includegraphics[width=0.9\columnwidth]{your-figure-name}
% \caption{Your figure caption here.}
% \label{fig:supp1}
% \end{figure}

% ----------- Supplementary Content Ends Here -----------

% References and End of Paper
% These lines must be placed at the end of your paper
%\bibliography{aaai2026}

\end{document}